# Enhancing the accuracies by performing pooling decisions adjacent to the output layer

Yuval Meir[1], Yarden Tzach[1], Ronit D. Gross[1], Ofek Tevet[1], Roni Vardi[2] and Ido Kanter[1,2,*]

[1]Department of Physics, Bar-Ilan University, Ramat-Gan, 52900, Israel.
[2]Gonda Interdisciplinary Brain Research Center, Bar-Ilan University, Ramat-Gan, 52900, Israel.

[*]Corresponding author email: ido.kanter@biu.ac.il

**Learning classification tasks of $(2^n \times 2^n)$ inputs typically consist of $\leq n$ $(2 \times 2)$ max-pooling (MP) operators along the entire feedforward deep architecture. Here we show, using the CIFAR-10 database, that pooling decisions adjacent to the last convolutional layer significantly enhance accuracies. In particular, average accuracies of the advanced-VGG with $m$ layers (A-VGGm) architectures are 0.936, 0.940, 0.954, 0.955, and 0.955 for m=6, 8, 14, 13, and 16, respectively. The results indicate A-VGG8's accuracy is superior to VGG16's, and that the accuracies of A-VGG13 and A-VGG16 are equal, and comparable to that of Wide-ResNet16. In addition, replacing the three fully connected (FC) layers with one FC layer, A-VGG6 and A-VGG14, or with several linear activation FC layers, yielded similar accuracies. These significantly enhanced accuracies stem from training the most influential input-output routes, in comparison to the inferior routes selected following multiple MP decisions along the deep architecture. In addition, accuracies are sensitive to the order of the non-commutative MP and average pooling operators adjacent to the output layer, varying the number and location of training routes. The results call for the reexamination of previously proposed deep architectures and their accuracies by utilizing the proposed pooling strategy adjacent to the output layer.**

**Introduction:**

Classification tasks are typically solved using deep feedforward architectures[1-6]. These architectures are based on consecutive convolutional layers (CLs) and terminate with a few fully connected (FC) layers, in which the output layer size is equal to the number of input object labels. The first CL functions as a filter revealing a local feature in the input, whereas consecutive CLs are expected to expose complex, large-scale features that finally characterize a class of inputs[1,7-10].

The deep learning (DL) strategy is efficient only if each CL consists of many parallel filters, the layer's depth, which differ by their initial convolutional weights. The depth typically increases along the deep architecture, resulting in enhanced accuracy. In addition, given a deep architecture and the ratios between the depths of consecutive CLs, accuracies increase as a function of the first CL depth.[11]

The deep learning strategy resulted in several practical difficulties, including the following. First, although the depth increases along the deep architecture, the input size of the layers remains fixed. The second difficulty is that the last CL output size, depth × layer input size, becomes very large, serving as the first FC layer input, which consists of a large number of tunable parameters. These computational complexities overload even powerful GPUs, limited by the accelerated utilization of a large number of filters and sizes of the FC layers. One way to circumvent these difficulties is to embed pooling layers along the CLs[1]. Each pooling reduces the output dimension of a CL by combining a cluster of outputs, e.g., $2 \times 2$, at one, and $n$ such operations along the deep architecture reduce the CL dimension by a factor $4^n$. The most popular pooling operators are max-pooling (MP)[12], which implements the maximal value of each cluster, and average pooling (AP)[13,14], which implements the average value of each cluster; however, more types of pooling operators exist [12,15-17].

The core question in this work is whether accuracies can be enhanced depending on the location of the pooling operators along the CLs of a given deep architecture. For instance, VGG16 consists of 13 CLs, three FC layers, and five (2 × 2) MP operators located along the CLs² (Fig. 1A). The results indicate that accuracies can be significantly increased by a smaller number of pooling operators adjacent to the last CL with optionally larger pooling sizes, for example, the advanced VGG16 (Fig. 1B). The optimized accuracies of these types of advanced VGG architectures with $m$ layers (A-VGGm) are first presented for selected $m$ values ($6 \leq m \leq 16$). Next, the underlying mechanism of the enhanced A-VGGm accuracies is discussed.

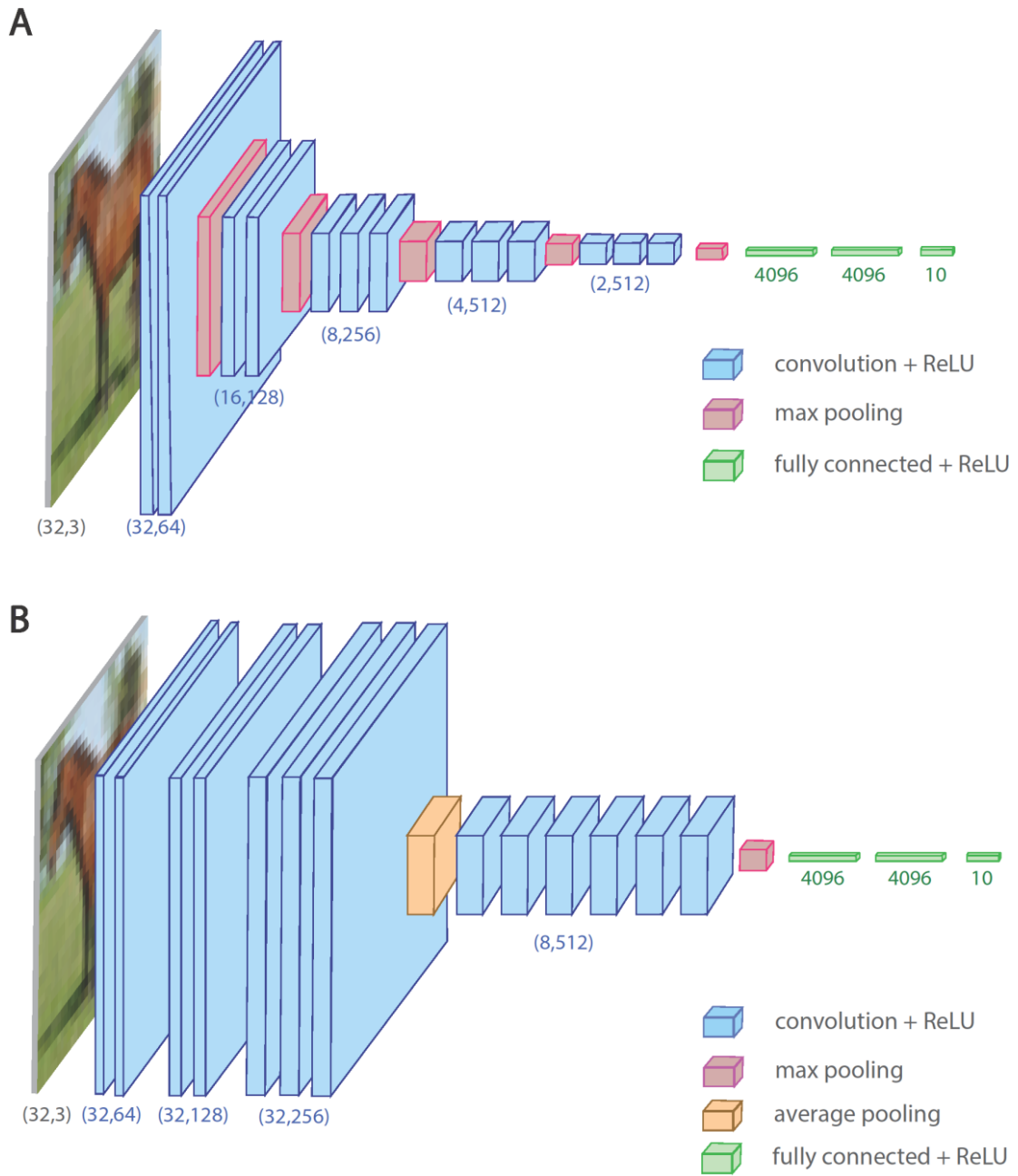

**Figure 1: VGG16 and A-VGG16 architectures. (A)** VGG16 architecture (13 $(3 \times 3)$ CLs and in between 5 $(2 \times 2)$ MP operators, followed by 3 FC layers) for the CIFAR10 database consisting of $32 \times 32$ RGB inputs. A CL is defined by its square filters with dimension K and depth D, $(K, D)$. **(B)** A-VGG16 architecture for CIFAR10 inputs consisting of 7 $(3 \times 3)$ CLs, $(4 \times 4)$ average pooling (AP), 6 $(3 \times 3)$ CLs, $(2 \times 2)$ MP and 3 FC layers.

**A-VGGm accuracies:**

A-VGG16 consists of $(4 \times 4)$ average pooling (AP) and $(2 \times 2)$ MP after the 7th and 13th CL, respectively (Fig. 1B and Table 1), with a maximal depth of 512 as in VGG16. The maximum average accuracy, 0.955, is superior to the accuracy, 0.935, obtained for the standard VGG16[11,18] and the fine-tuned optimized accuracy 0.94[10] (Fig. 1A) and is comparable with the Wide-ResNet[4,5] median accuracy consisting of 16 layers with widening factor 10 (WRN16-10).

Note that the replacement of the pair of pooling operators, $[AP(4 \times 4), MP(2 \times 2)]$ along A-VGG16 (Fig. 1B), by several other options, for example, $[MP(2 \times 2), AP(4 \times 4)]$ and $[AP(2 \times 2), MP(8 \times 8)]$, also yielded an average accuracies $> 0.95$, indicating the superior robustness of A-VGG16 accuracies over VGG16. Removing the last three CLs (the fifth block of A-VGG16) resulting in A-VGG13, with an average accuracy of 0.955, identical to that of A-VGG16 up to the first three leading digits (Table 1). One possible explanation to the same accuracies is that the receptive field[19] of the last three CLs of A-VGG13 is $7 \times 7$ saturating the $8 \times 8$ layers' input size. It also suggests that accuracies are only mildly affected by $m > 13$.

The A-VGG8 architecture that consists of only 8 layers, results in 0.940 averaged accuracy, exceeding the optimized VGG8 accuracy of 0.915, which consists of 5 $MP(2 \times 2)$ one after each CL[2,20], and also exceeds the average accuracy of VGG16. Here again, $AP(2 \times 2)$ and $MP(4 \times 4)$ were placed after the 3rd and the 5th CLs, respectively (Table 1). This result indicates that a shallow architecture, with fewer pooling operators adjacent to the output, can imitate the accuracies of a deeper architecture with double the number of layers, while the receptive field covers a small portion of the layers' input size.

Using only one FC layer reduces the number of layers by two, from A-VGG16 to A-VGG14, and from A-VGG8 to A-VGG6 (Table 1). The results indicate that accuracies are only mildly affected by such modifications, where A-VGG6 achieves an average accuracy of 0.936, which slightly exceeds that of VGG16 and A-VGG14 exceeds 0.954 (Table 1). We note that this type of architectures with only one FC layer consists of fewer parameters and can be mapped onto tree architectures[21].

Similarly, the A-VGG13 and A-VGG16 architectures with linear activation functions for the FC layers achieved similar averaged accuracies of $0.954$ and $0.955$, respectively, both with small standard deviations (Supplementary Information). The three linear FC layers can be folded into one in the test procedure[22], minimizing its latency; however, training must be performed with three separated FC layers.

The gap between the average accuracies of A-VGG8 and A-VGG6 (~0.004) was slightly greater than that between the enhanced accuracies of A-VGG16 and A-VGG14 (Table 1), indicating that the gap decreases with $m$. Nevertheless, the comparable average accuracies of A-VGG13 and A-VGG14 with A-VGG16 indicate that removing two out of three FC layers or removing three out of the thirteen CLs does not affect accuracies. Hence, it is interesting to examine the average accuracies of VGG11 where two FC layers as well as the last three CLs are removed.

| A-VGG6 | A-VGG8 | A-VGG13 | A-VGG14 | A-VGG16 |
|---|---|---|---|---|
| Conv.1-64 | Conv.1-64 | Conv.2-64 | Conv.2-64 | Conv.2-64 |
| Conv.1-128 | Conv.1-128 | Conv.2-128 | Conv.2-128 | Conv.2-128 |
| Conv.1-256 | Conv.1-256 | Conv.3-256 | Conv.3-256 | Conv.3-256 |
| AP 2x2 | AP 2x2 | AP 4x4 | AP 4x4 | AP 4x4 |
| Conv.2-512 | Conv.2-512 | Conv.3-512 | Conv.6-512 | Conv.6-512 |
| MP 8x8 | MP 4x4 | MP 4x4 | MP 2x2 | MP 2x2 |
| FC x10 | FC 8192 | FC x2048 | FC x10 | FC x4096 |
|  | FC 8192 | FC x2048 |  | FC x4096 |
|  | FC x10 | FC x10 |  | FC x10 |
| **Avg. Accuracy** | | | | |
| 0.936 | 0.940 | 0.955 | 0.954 | 0.955 |

**Table 1: Architectures and accuracies of A-VGGm.** A-VGGm architectures, m=6, 8, 13, 14, and 16, and their maximized average accuracies obtained from 10 samples (detailed parameters and accuracies' standard deviations are presented in the Supplementary Information).

**Optimized learning gain using pooling operators:**

The backpropagation learning step[23] updates the weights towards the correct output values for a given input. Typically, such a learning step can add noise and is destructive to a fraction of the training set[24-27]. However, the average accuracy increases with epochs and asymptotically saturates at a value that identifies the quality of the learning algorithm for a given architecture and database.

One important ingredient of DL is downsizing the input size as the layers progress. This can be done by either pooling operators or using the stride of the CLs. Although both reduce the size of the input, the pooling operators transfer specific output fields, such as maximal field in the MP operators. It aims to select the most influential field from a small cluster on a node in the successive layer, for example, MP ($2 \times 2$). Its underlying logic is to maximize the learning step gain for the current input while minimizing the added noise by zeroing other routes; *maximize learning with minimal side-effect damage*. However, this local maximization does not ensure a global one.

Commonly, several MP operators are placed among the CLs, for example, five times in the case of VGG16 (Fig. 1A), and apparently solve simultaneously the following two difficulties. First, although the depth, $D$, increases along the CLs (Fig. 1A), the input size, $K$, of the layers shrinks accordingly such that the output sizes of the CLs, $K \times D$, do not grow linearly with depth. Second, successive MP operators appear to select the most influential routes on the first FC layer, which is adjacent to the output layer. However, these local decisions following consecutive MP operators do not necessarily result in the most influential routes in the first FC layer, as elaborated below using a toy model.

Assume a binary tree, where its random nodal values are low, medium, or high (Fig. 2A). The tree output is equal to the branch with the maximal field, which is calculated as the product of its three nodal values. The first strategy is based on local decisions, similar to MP operators. For each node the maximal child, among the two, is selected (gray circles in Fig. 2A), and the selected route is the one composed of gray nodes only, where its value is $M \cdot M \cdot M$ (the brown branch in Fig. 2A). However, a global decision among the eight branches results in a maximal field $H \cdot H \cdot L$ (green branch in Fig. 2A). This toy model indicates that a global decision differs from local decisions; however, the probability of such an event is unclear.

A more realistic model, imitating deep architectures (Fig. 1), is Gaussian random ($1024 \times 1024$) inputs followed by ten ($3 \times 3$) CLs with unity depth (Fig. 2B). Two scenarios, local decisions and a global decision, are discussed. In the first, ($2 \times 2$) MP operators are placed after each of the first $n$ CLs (Fig. 2B top, exemplified $n = 4$), where in the second one a single ($2^n \times 2^n$) MP operator is placed after the ten CLs (Fig. 2B bottom, exemplified $n = 4$). For both scenarios, there are ($2^{10-n} \times 2^{10-n}$) non-negative (ReLU) outputs, denoted by $O_{SP}$ (Sequence Pooling), representing local decisions and $O_{TP}$ (Top Pooling), representing a global decision. For a given $n$, the $2^{10-n} \times 2^{10-n}$ ratios, $O_{SP}/O_{TP}$, were calculated and averaged over many Gaussian random inputs and several sets of ten randomly selected convolutional filters, which were identical for both scenarios. The probability $P(\frac{O_{SP}}{O_{TP}} > 1)$ indicates that local decisions, $n$ consecutive MPs, result in a larger output than a global decision, a single ($2^n \times 2^n$) MP (Supplementary Information). This probability rapidly decreased with $n$, possibly exponentially (Fig. 2C), and even for $n = 2$ it was below $0.1$.

The increase in CLs depth beyond unity does not qualitatively affect the probability $P\left(\frac{O_{SP}}{O_{TP}} > 1\right)$, as indicated by simulations of VGG8 with five consecutive $(2 \times 2)$ MP operators after each CL and a single $(32 \times 32)$ MP after five CLs. The same five random $(3 \times 3)$ convolutions were used for both architectures, and the 512 ratios, $\frac{O_{SP}}{O_{TP}}$ for the single output of each filter, were calculated. Averaging over CIFAR10 training inputs and several selected sets of fixed random convolutions results in $O(10^{-3})$ for probability $P(\frac{O_{SP}}{O_{TP}} > 1)$.

The results clearly indicate that a global decision selects the most influential routes to the first FC layer. Hence, pooling adjacent to the output layer, is superior to the selection following consecutive local pooling decisions. This supports that using larger pooling operators adjacent to the output of the CLs enhances accuracies (Table 1). It is expected that using pooling operators solely after the entire CLs will enhance accuracies even further; however, its validation in simulations of A-VGGm architectures is difficult. The running time per epoch of such large $K \times D$ deep architectures is several times longer, and the optimization of accuracies is currently beyond our computational capabilities.

A simpler architecture is the LeNet5[28,29], with much lower depth and total number of CLs, consisting of two CLs followed by $(2 \times 2)$ MP each and three FC layers (Fig. 3A). The optimized average accuracy on the CIFAR10 database is 0.77.[11] Advanced LeNet5 (A-LeNet5) architectures consist of pooling operators only after the second CL (Fig. 3A). In particular, the two pooling options, $(2 \times 2)\, AP \circ (2 \times 2)\, MP$ and $(2 \times 2)\, MP \circ (2 \times 2)\, AP$ were examined (examples $a$ and $b$ in Fig. 3A), imitating the dimensions of the two $(2 \times 2)\, MP$ of LeNet5. Indeed, these A-LeNet5 architectures enhance average accuracies by up to ~0.02, in comparison to LeNet5 (Fig. 3B), as predicted by the abovementioned argument. Similarly, using either $(4 \times 4)\, MP$ or $(4 \times 4)\, AP$ after the second FC layer resulted in ~0.79 maximized average accuracies (not shown). The shift of the MP by only one CL, from the first to the second, improves the accuracies, and an enhanced effect might be expected by skipping over more CLs in deeper architectures (Fig. 2C). An interesting aspect of A-LeNet5 is that accuracies improved although the receptive field covers only a small portion of the input, in contrast to A-VGG16.

Another type of A-LeNet5 is a combination of a pair of $(2 \times 2)$ and $(3 \times 3)$ pooling operators after the two CLs ($c$, $d$ and $e$ in Fig. 3A). Although the input size of the first FC layer decreased from $400$ in LeNet5 to $256$, the average accuracies was enhanced by $\sim 0.011$ ($d$ in Fig. 3B). This result exemplifies the improved A-LeNet5 accuracies even when the input size of the first FC layer decreases. Examples $d$ and $e$ (Fig. 3A) consist of the same pooling operators, $(2 \times 2)\ AP$ and $(3 \times 3)\ MP$, but with the exchanged order of the operators. Their average accuracies differ by $\sim 0.016$ (Fig. 3B), indicating that these pooling operators do not commute with the exchanged order. Another possible class of commutation is the exchanged type of operation (color) while maintaining their size; exchanged $MP$ and $AP$ ($c$ and $d$ or $a$ and $b$ in Fig. 3A). Average accuracies indicate that pooling operators do not necessarily commute with exchanged colors.

The two non-commutative classes, order and type of operations, stem from different numbers and locations of the backpropagation active routes in the lower layers (Fig. 3C). The number of locally active backpropagation routes in a $(6 \times 6)$ window is $9$ for $(3 \times 3)\ AP \circ (2 \times 2)\ MP$, whereas for $(3 \times 3)\ MP \circ (2 \times 2) AP$ is $4$. For the exchanged order of operators ($d$ and $e$ in Fig. 3A), the number of backpropagation active routes is the same, $4$, in both cases. However, these $4$ routes were localized in $(2 \times 2)$ ($e$ in Fig. 3B and Fig. 3C), but delocalized over $(6 \times 6)$ ($d$ in Fig. 3B and Fig. 3C). Hence, the non-commutation of pooling operators can stem either from the different numbers of active backpropagation routes or from their different locations.

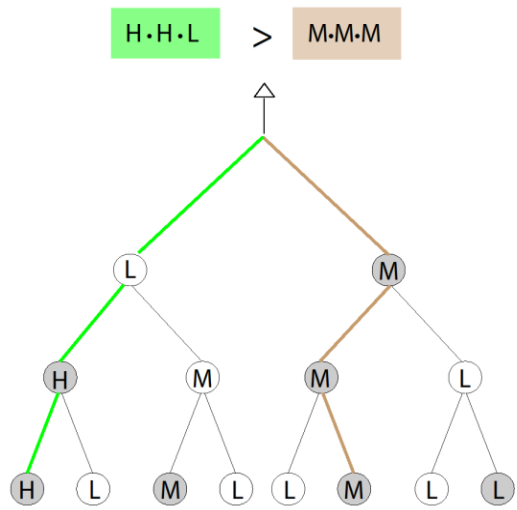
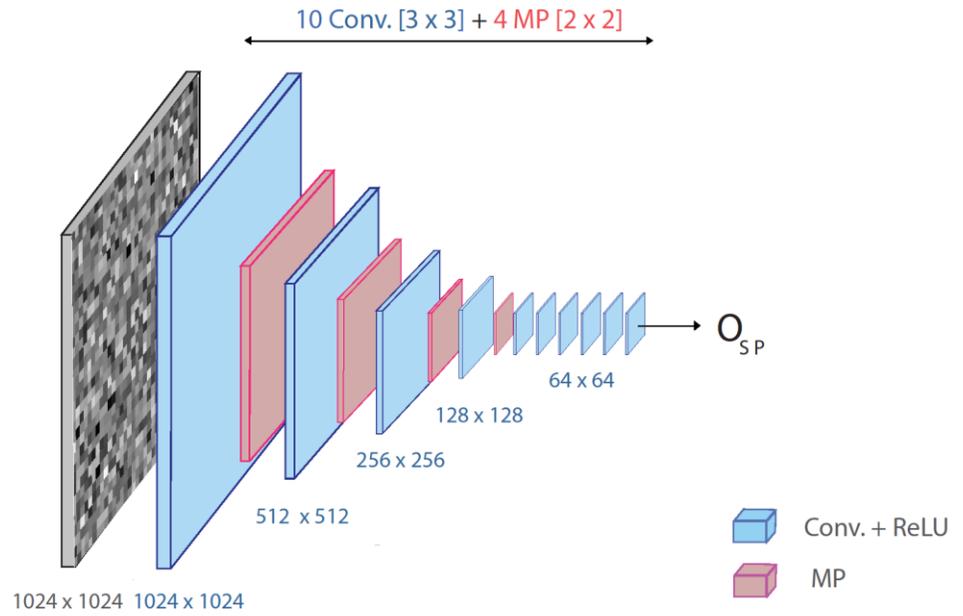
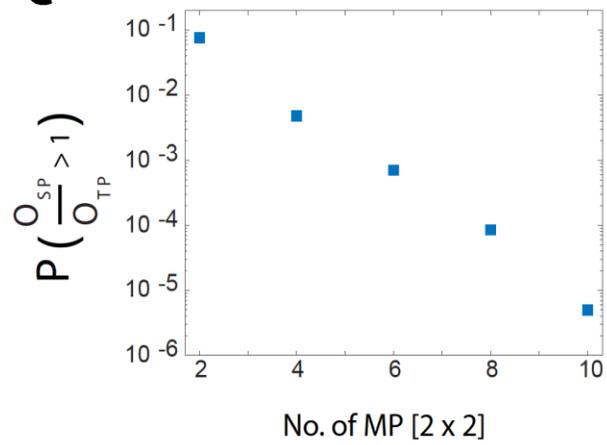
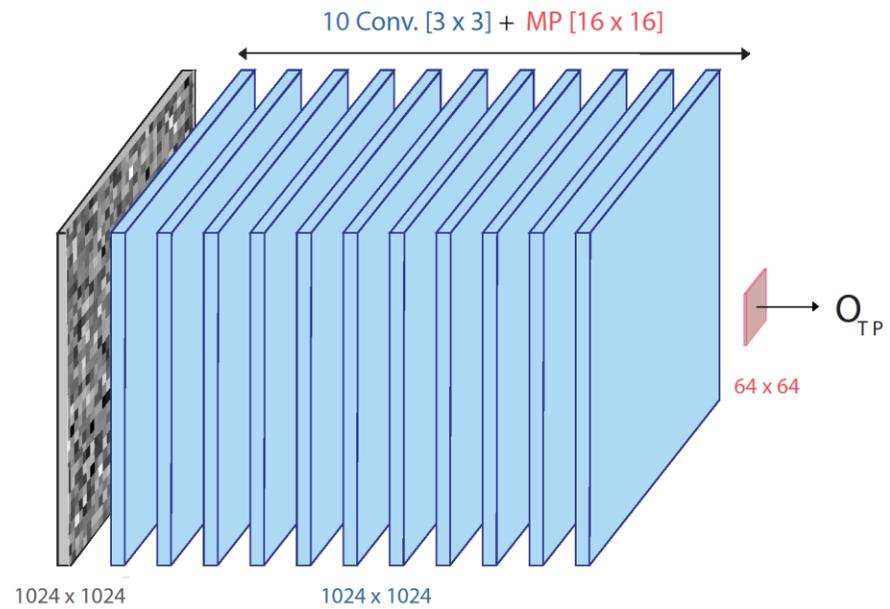

**Figure 2: Comparison between several small MP operators along CLs and a large one at their end.** **(A)** A binary tree where the random nodal values are low (L), medium (M), or high (H), e.g., $1, 10$, and $1000$. A local decision selects the path to the maximal nodal child (gray), resulting in the brown route connecting three gray nodes. A global decision selects the green route, maximizing the product of its nodal values. **(B)** Gaussian random $(1024 \times 1024)$ input followed by ten $(3 \times 3)$ CLs, where $(2 \times 2)$ MP is placed after the first four CLs (Top), and similar architecture where a single $(16 \times 16)$ MP is placed after the 10 CLs. The $(64 \times 64)$ output values are denoted by $O_{SP}$ (Top) and $O_{TP}$ (bottom) (Supplementary Information). **(C)** The probability $P(\frac{O_{SP}}{O_{TP}} > 1)$ as a function of the number, $n$, of $(2 \times 2)$ MP ($n = 4$ is demonstrated in **B**).

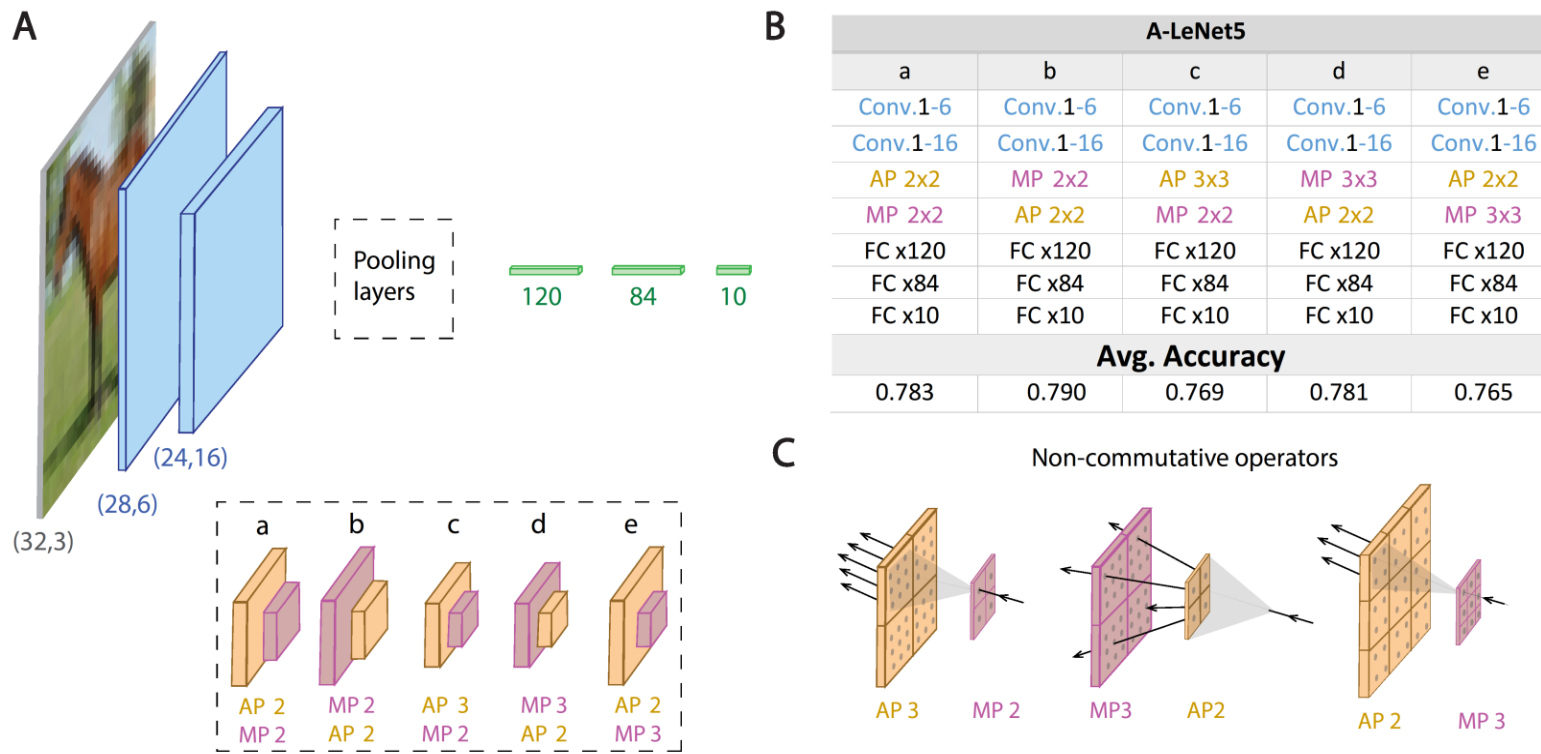

**Figure 3: A-LeNet5 accuracies' architectures and the role of non-commutative pooling operators. (A)** A-LeNet5 architectures, where the pooling layers (exemplified in the dashed rectangle) are placed after the second CL. **(B)** Detailed architectures and average accuracies for the five schemes of A-LeNet5 in **A** (see Supplementary Information for detailed parameters and standard deviations). **(C)** Non-commutative pooling operators, where the number of active backpropagation routes is 9 for $(3 \times 3)\, AP \circ (2 \times 2)\, MP$, 4 delocalized routes for $(3 \times 3)\, MP \circ (2 \times 2)\, AP$ and 4 localized routes for $(2 \times 2)\, AP \circ (3 \times 3)\, MP$.

**Discussion:**

The aim of pooling operators, is downsizing the input size as the layers progress while transferring specific output fields, such as maximal field in the MP operators. It selects the most influential local field, but does not ensure a most influential global field on the output. The proposed enhanced learning strategy is based on updating the most influential routes, that is, the maximal fields, on the output units. This is supported by the A-VGGm and A-LeNet5 simulations, where the average accuracies are enhanced using pooling operators placed closer to the output layer (Fig. 1, Table 1, and Fig. 3). Its underlying mechanism is aimed at maximizing the learning gain for the current input, while simultaneously minimizing the average damage on the current learning of the entire training set. Each learning step for a given input induces noise on the learning of other inputs. Hence, increasing the signal-to-noise ratio (SNR) of a learning step, average over the training set, requires updating the most influential routes of the current input; *maximize learning with minimal side-effect damage*.

The realization of the proposed advanced learning strategy entails a discussion of the following three theoretical and practical aspects. First, the selection of the most influential routes on the first FC layer is not necessarily equivalent to the selection of the most influential routes on the output units. However, a backpropagation step initiated at the most influential input weight on an output unit, updates all the CLs' routes since the spatial structure disappears within the one-dimensional FC layers. Hence, the proposed strategy approximates only the most influential routes on the outputs. The exceptional architectures were A-VGG6 and A-VGG14 (Table 1), consisting of one FC layer, demonstrating accuracies that were only slightly below A-VGG8 and A-VGG16, respectively.

The second aspect concerns the computational complexity of the proposed advanced learning strategy. Selecting the most influential routes after all CLs with their fixed depth overloaded even advanced GPUs since the depth increases while the layer's dimension does not decrease. For instance, the running time per epoch of $(32 \times 32)$ MP placed after all CLs of A-VGG16 was slowed down by a factor of $\sim 10$. To circumvent this difficulty, the advanced learning strategy was approximated by placing the first pooling operator before the CLs with maximal depth and the second operator after all CLs (Fig. 1 and Table 1). Nevertheless, it is interesting to examine, using advanced GPUs, whether placing pooling operators after all CLs further advances accuracies.

The third aspect is the selection of the types, dimensions, and locations of pooling operators along the deep architecture to maximize accuracies. For a given A-VGGm, several pooling arrangements result in similar accuracies, and we report only the one that maximizes the average accuracies under a given number of epochs. Nevertheless, the maximized average A-VGGm accuracies hint at preferred combinations where the AP is placed before CLs with maximal depth and the MP operates after all CLs (Table 1 and Fig. 1), which might stem from the following insight. MP after all CLs carefully selects only one significant backpropagation route among a cluster of routes, whereas an AP close to the input layer spreads its incoming backpropagation signal to multiple routes. This arrangement was found to maximize accuracies for several A-VGGm architectures (Table 1). However, A-LeNet indicated an opposite trend, where AP at the top of two adjacent pooling operators maximized accuracies (Fig. 3). The role is not yet clear and may depend on the database and details of the training architecture.

We present an argument indicating that pooling decision adjacent to the output layer enhances accuracy (Table 1). However, one might attribute this improvement to the increase in the number of parameters in the first FC layer, where the number of parameters in the rest of CLs and FC layers remain the same. In order to pinpoint the accuracy improvement to the location of the pooling operators, we obtained ~0.954 for A-VGG16 with $4 \times 4$ AP after the 7[th] CL and with $8 \times 8$ MP operator after the 13[th] CL. In this architecture the size of the first FC layer is the same as in VGG16, and therefore the number of parameters in both remain the same, yet there is a clear improvement in the accuracy.

The non-commutative pooling operator features exemplify the sensitivity of the maximal average accuracies to their order and type, and significantly enrich the possible number of pooling operators with a given dimension. For $(8 \times 8)$ pooling dimension, for instance, one can find 8 possible pooling operators; $(2 \times 2)XP \circ (2 \times 2)YP \circ (2 \times 2)ZP$, where $X, Y$ and $Z$ equal either to $M$ (Max) or $A$ (Average). Similarly, the number of pooling operators with dimension $(2^n \times 2^n)$ is $2^n$, and exponentially increases when more than two types of $(2 \times 2)$ pooling operators are allowed. The results for A-LeNet indicate that enhanced accuracies can be achieved using combinations of consecutive pooling operators after the second CLs (Fig. 3). However, the identification of preferred combinations to maximize the accuracies in general deep architectures deserves further research.

The non-commutative features of pooling operators also stem from their different number of backpropagation downstream updated routes (Fig. 3C). For instance, A-VGG16 with $(32 \times 32)$ MP, before the first FC layer, consists of a single backpropagation downstream updated route per filter, whereas for $(32 \times 32)$ AP there are $1024$ routes. Nevertheless, the preferred pooling operators to maximize accuracies need to be determined. The most influential route is favored to correct the output of the current input; however, it induces output noise on other training inputs, resulting in a low SNR. Similarly, updating $1024$ downstream routes using AP, including the weak ones, increases the correct output of the current input in comparison to MP; however, with enhanced side-effect, noise on other training inputs, resulting in a possibly low SNR. Hence, for a given architecture and dataset, the selection of pooling operators that maximize the averaged SNR per epoch is yet unclear.

The accuracies of A-VGG6 and A-VGG14 with only one FC layer were only slightly below those of the three FC layers, A-VGG8 and A-VGG16, respectively (Table 1). Architectures with only one FC layer are characterized by lower learning complexity and number of tunable parameters. In addition, these architectures can be mapped onto tree architectures[30,31], generalizing recent LeNet mapping into tree architecture without affecting accuracies but with lower computational learning complexity[31]. Tree mapping of architectures comprising more than two CLs, inspired by dendritic tree learning[30-35], is beyond the scope of the presented work and will be discussed elsewhere.

It was observed that shallower architectures can yield the same accuracies as deeper ones, for instance, A-VGG13 and A-VGG16. This result can be attributed to fact that the last CLs' receptive field completely covers the input, suggesting that the last three CLs are redundant. Another example, is A-VGG8 which achieves the same accuracy as VGG16. In this case the last CLs of A-VGG8, the receptive field does not fully cover the CLs' input. Hence, the enhanced accuracy is attributed to the advanced location of the pooling operators. Similarly, A-LeNet5 enhances the accuracies of LeNet5, while the receptive field covers a small portion of the input, in contrast to A-VGG16.

The extension of the proposed idea to deeper architectures on CIFAR-10, e.g. DenseNet[36] and EfficientNet[37], results in the following several difficulties which at the moment are beyond our computational capabilities. The accuracy of deeper architectures approaches one and thus the enhancement of the accuracy by preforming pooling decisions adjacent to the output is expected to be in a sub-percentage increase. The observation of such minor accuracy improvements will require fine-tuned optimization in high resolution on the hyper-parameter space as well as large statistics. We note that the running time per epoch even for A-VGG16 where a $(32 \times 32)$ $MP$ was placed after all CLs was slowed down by a factor of $\sim 10$, which made its optimization beyond our computational capabilities.

Using datasets with higher complexity, more classes and a lesser number of training examples per class, e.g., CIFAR-100[38] and ImageNet[39], result in significant fluctuated accuracy among samples. These fluctuations make observing the effect of the pooling operators adjacent to the output layers much more difficult, and deserves careful further examination using more advanced computational capabilities.

The original VGG architectures were constructed for large input image sizes of $224 \times 224$. The presented work demonstrates enhanced accuracies using A-VGGm architectures on small input image sizes, $32 \times 32$. Extrapolating these enhancements on large images is much beyond our computational capabilities. Nevertheless, preliminary results using online learning (one epoch) on images of size $128 \times 128$ indicate a slight improvement of the average accuracies of

A-VGG16 in comparison to VGG16. In general one might expect that the kernel size in A-VGGm might require scaling with the size of the input images in order to have the entire input covered by the receptive field.

Finally, the reported average accuracies for A-VGG16 and A-VGG13 approach Wide-ResNet16 (widening factor of 10) median accuracies, consisting of an architecture with three main ingredients: skip connections, CLs with stride $= 2$ and $(8 \times 8)$ AP after all CLs. This similarity hints that the ingredient dominating the enhanced accuracies, among the three, is a pooling operation after all CLs.

## Data availability

Source data were provided in this study. All data supporting the plots within this paper, along with other findings of this study, are available from the corresponding author upon reasonable request.

## Code availability

The simulation code is provided in this study, parallel to its publication in GitHub.

## Acknowledgments

We thank for the much helpful suggestions and insights of both reviewers which helped improving the final version of this paper. I.K. acknowledges the partial financial support from the Israel Science Foundation (grant number 346/22).

## Author contributions

Y.M., Y.T., R. D. G., and O.T. contributed equally to this study. R. V. discussed the results and commented on the manuscript. I.K. initiated and supervised all aspects of the study.

## Competing interests

The authors declare no competing interests.

## Additional information

Supplementary information. The online version contains supplementary material available at

# Supplementary Information

**Enhancing the accuracies by performing pooling decisions adjacent to the output layer**

Yuval Meir[1], Yarden Tzach[1], Ronit D. Gross[1], Ofek Tevet[1], Roni Vardi[2] and Ido Kanter[1,2,*]

[1]Department of Physics, Bar-Ilan University, Ramat-Gan, 52900, Israel.
[2]Gonda Interdisciplinary Brain Research Center, Bar-Ilan University, Ramat-Gan, 52900, Israel.

[*]Corresponding author email: ido.kanter@biu.ac.il

**Advanced VGGm architectures.** The examined advanced VGGm (A-VGGm) architectures consist of m layers, $6 \leq m \leq 16$ (Fig. 1A-B, exemplifies $m = 16$ for VGG16[1] and A-VGG16.

For $m = 6$ and 8, the architecture is similar to the VGG8[1], with initial depth of 64 for the first CL and doubling depth for the next three CLs, and with a single zero-padding around the input of each CL. For $m = 6$, a $(2 \times 2)$ average pooling (AP) is applied after the third CL and an $(8 \times 8)$ max-pooling (MP) after the fifth CL. For $m = 8$, a $(2 \times 2)$ AP is applied after the third CL and a $(4 \times 4)$ MP after the fifth CL. $m = 6$ terminates with one FC layer consisting of 2048 hidden units and $m = 8$ with three FC layers with 8192 hidden units each.

For $m = 14$ and 16, there are 13 CLs with doubling depth (except for the last 3 CLs) and with a single zero-padding around the input of each CL, followed by one FC layer with 8192 hidden units for $m = 14$ and three FC layers with 4096 hidden units for $m = 16$. For both $m = 14$ and 16, a $(4 \times 4)$ AP is applied after the 7th CL and a $(2 \times 2)$ MP is applied after the 13th CL.

For $m = 13$ the last three CLs are withdrawn, resulting in ten CLs, where a $(4 \times 4)$ AP is also applied after the 7th CL and a $(4 \times 4)$ MP is applied after the 10th CL terminating with three FC layers consisting of 2048 hidden units each.

After each CL, a batch normalization layer was applied. The softmax function was applied to the ten outputs. The ReLU activation function was assigned to each hidden unit (not including the ten output units and pooling operators), and all weights were initialized using a uniform distribution with a zero mean and unity standard deviation (Std) according to the He normal initialization[3].

For A-VGG13 and A-VGG16 with linear activation functions for the FC layer the architectures remain the same.

**Advanced LeNet5 architectures.** The advanced LeNet5 (A-LeNet5) architectures consist of two consecutive CLs of size $(5 \times 5)$ with depths $d_1 = 6$ and $d_2 = 16$ and three FC layers (Fig. 3A). These architectures are similar to the LeNet5[2], however, the pooling operators are applied only after the second CL (Fig. 3A). The ReLU activation function was assigned to each hidden unit where the softmax function was applied to the ten output units. All weights were initialized using a uniform distribution with a zero mean and unity Std according to the He normal initialization[3].

**Data preprocessing.** Each input pixel of an image ($32 \times 32$) from the CIFAR-10 database was divided by the maximal pixel value, 255, multiplied by 2, and subtracted by 1, such that its range was $[-1, 1]$. In all simulations, data augmentation was used, derived from the original images, by random horizontally flipping and translating up to four pixels in each direction.

**Optimization.** The cross-entropy cost function was selected for the classification task and was minimized using the stochastic gradient descent algorithm[4,5]. The maximal accuracy was determined by searching through the hyper-parameters (see below). Cross-validation was confirmed using several validation databases, each consisting of 10,000 random examples from the training set, as in the test set. The averaged results were in the same Std as the reported average success rates. The Nesterov momentum[3] and L2 regularization method[4] were applied.

**Hyper-parameters.** The hyper-parameters η (learning rate), μ (momentum constant[3]), and α (regularization L2[4]) were optimized for offline learning, using a mini-batch size of 100 inputs. The learning rate decay schedule[5,6] was also optimized such that it was multiplied by the decay factor, q, every $\Delta t$ epochs, and is denoted below as $(q, \Delta t)$.

**Out of phase scheduling.** For A-VGG16 and A-VGG8 the decay schedules[5,6] of the FC layers and the CLs learning rates had a phase of 10 epochs in between. The decay scheduling for the FC layers starts at $epoch = 10$, while for the CLs at $epoch = 20$. Specifically, decay learning rate of the FC layers occurs at $epochs = [10, 30, 50 \ldots]$, while for the CLs at $epochs = [20, 40, 60 \ldots]$.

**Fig 1. A-VGGm Hyper-parameters**

| A-VGG16 | | | | |
|---|---|---|---|---|
| Layer | η | μ | α | epochs |
| CLs | 0.00721 | 0.98 | 1.15e-3 | 280 |
| FC layers | 0.0045 | 0.982 | 1.35e-3 | 280 |

The decay schedule for the learning rate is defined as follows:

For CLs:

$$(q, \Delta t) = \begin{cases} (0.65, 20) & \text{epoch} \leq 140 \\ (0.55, 20) & \text{epoch} > 140 \end{cases}$$

For FC layers, with 10 epochs out of phase:

$$(q, \Delta t) = \begin{cases} (0.65, 20) & \text{epoch} < 150 \\ (0.5, 20) & \text{epoch} \geq 150 \end{cases}$$

The accuracies' Std is 0.0015.

| A-VGG14 | | | | |
|---|---|---|---|---|
| Layers | η | μ | α | epochs |
| CLs | 0.0078 | 0.985 | 1.15e-3 | 200 |
| FC layers | 6.05e-4 | 0.98 | 1.15e-3 | 200 |

The decay schedule for the learning rate is defined as follows:

For CLs:

$$(q, \Delta t) = (0.65, 20)$$

For FC layers:

$$(q, \Delta t) = \begin{cases} (0.55, 10) & \text{epoch} < 120 \\ (0.5, 10) & \text{epoch} \geq 120 \end{cases}$$

The accuracies' Std is 0.00092.

| A-VGG13 | | | | |
|---|---|---|---|---|
| Layers | η | μ | α | epochs |
| CLs | 0.0078 | 0.98 | 1.15e-3 | 200 |
| FC layers | 0.00297 | 0.985 | 1.15e-3 | 200 |

The decay schedule for the learning rate is defined as follows:

For CLs:

$$(q, \Delta t) = (0.65, 20)$$

For FC layers:

$$(q, \Delta t) = \begin{cases} (0.55, 20) & \text{epoch} < 120 \\ (0.5, 20) & \text{epoch} \geq 120 \end{cases}$$

The accuracies' Std is 0.0012.

| A-VGG8 | | | | |
|---|---|---|---|---|
| Layers | $\eta$ | $\mu$ | $\alpha$ | epochs |
| CLs | 0.0145 | 0.97 | 1e-3 | 200 |
| FC layers | 0.002 | 0.975 | 1.2e-3 | 200 |

The decay schedule for the learning rate is defined as follows:

For CLs:

$$(q, \Delta t) = \begin{cases} (0.66, 20) & \text{epoch} \leq 140 \\ (0.55, 20) & \text{epoch} > 140 \end{cases}$$

For FC layers with 10 epochs out of phase:

$$(q, \Delta t) = \begin{cases} (0.66, 20) & \text{epoch} < 150 \\ (0.5, 20) & \text{epoch} \geq 150 \end{cases}$$

The accuracies' Std is 0.0009.

| A-VGG6 | | | | |
|---|---|---|---|---|
| Layers | $\eta$ | $\mu$ | $\alpha$ | epochs |
| CLs | 9.75e-3 | 0.972 | 1.1e-3 | 200 |
| FC layers | 1.95e-3 | 0.98 | 1.1e-3 | 200 |

The decay schedule for the learning rate is defined as follows:

For CLs:

$$(q, \Delta t) = \begin{cases} (0.65, 20) & \text{epoch} < 120 \\ (0.55, 20) & \text{epoch} \geq 120 \end{cases}$$

For FC layers:

$$(q, \Delta t) = \begin{cases} (0.65, 20) & \text{epoch} < 120 \\ (0.5, 20) & \text{epoch} \geq 120 \end{cases}$$

The accuracies' Std is 0.00224.

**A-VGG13 and A-VGG16 with linear activation functions for the FC layers.**

| A-VGG16 with linear activation | | | | |
|---|---|---|---|---|
| Layer | η | μ | α | epochs |
| CLs | 0.0078 | 0.98 | 1.15e-3 | 200 |
| FC layers | 0.00297 | 0.985 | 1.15e-3 | 200 |

The decay schedule for the learning rate is defined as follows:

For CLs:

$$(q, \Delta t) = (0.65, 20)$$

For FC layers:

$$(q, \Delta t) = \begin{cases} (0.55, 20) & \text{epoch} < 120 \\ (0.5, 20) & \text{epoch} \geq 120 \end{cases}$$

The accuracies' Std is 0.0014.

| A-VGG13 with linear activation | | | | |
|---|---|---|---|---|
| Layer | η | μ | α | epochs |
| CLs | 0.0078 | 0.98 | 1.15e-3 | 200 |
| FC layers | 0.00297 | 0.985 | 1.15e-3 | 200 |

The decay schedule for the learning rate is defined as follows:

For CLs:

$$(q, \Delta t) = (0.65, 20)$$

For FC layers:

$$(q, \Delta t) = \begin{cases} (0.55, 20) & \text{epoch} < 120 \\ (0.5, 20) & \text{epoch} \geq 120 \end{cases}$$

The accuracies' Std is 0.001.

**Fig. 2.** In Fig. 2, two architectures were compared, each consists of ten CLs with unity depth and the same ten $(3 \times 3)$ filters. Random inputs of size $(1024 \times 1024)$ with values taken from a Gaussian distribution with zero mean and unity Std were tested. The ReLU activation function was assigned to all the hidden and output units.

Two architectures were compared, sequence pooling (SP) and top pooling (TP). The SP architecture consists of $(2 \times 2)$ MP after each one of the first $n$ CLs, whereas for the TP architecture $(2^n \times 2^n)$ MP is applied after the 10$^{th}$ CL (Fig. 2B top, exemplifies $n = 4$). For a given $n = [2, 4, 6, 8, 10]$, the $(2^{10-n} \times 2^{10-n})$ SP and TP output ratios, $\frac{O_{SP}}{O_{TP}}$, were calculated and the probability $P\left(\frac{O_{SP}}{O_{TP}} > 1\right)$ was estimated using 20,000 - 100,000 different random inputs (depending on $n$), and filters which were randomly initialized for the CLs on each sample.

The increase in CLs depth beyond unity does not qualitatively affect $P\left(\frac{O_{SP}}{O_{TP}} > 1\right)$, as indicated by simulations of VGG8 with five consecutive $(2 \times 2)$ MP operators after each CL and a single $(32 \times 32)$ MP after five CLs. The same five random $(3 \times 3)$ convolutions were used for both architectures, and the 512 ratios, $\frac{O_{SP}}{O_{TP}}$, for the single output of each filter, were calculated. The probability $P\left(\frac{O_{SP}}{O_{TP}} > 1\right)$ was calculated by averaging over randomly selected batches of CIFAR10 training inputs and several selected sets of fixed random convolutions.

**Fig 3. A-LeNet5 Hyper-parameters**

| A-LeNet5 - a | | | |
|---|---|---|---|
| η | μ | α | epochs |
| 0.032 | 0.92 | 5e-4 | 240 |

The decay schedule for the learning rate is defined as follows:

$$(q, \Delta t) = \begin{cases} (0.8, 10) & \text{epoch} < 120 \\ (0.7, 10) & \text{epoch} \geq 120 \end{cases}$$

The accuracies' Std is 0.0028.

| A-LeNet5 - b | | | |
|---|---|---|---|
| $\eta$ | $\mu$ | $\alpha$ | epochs |
| 0.03 | 0.93 | 4e-4 | 280 |

The decay schedule for the learning rate is defined as follows:

$$(q, \Delta t) = \begin{cases} (0.8, 10) & \text{epoch} < 120 \\ (0.7, 10) & \text{epoch} \geq 120 \end{cases}$$

The accuracies' Std is 0.003.

| A-LeNet5 - c | | | |
|---|---|---|---|
| $\eta$ | $\mu$ | $\alpha$ | epochs |
| 0.028 | 0.925 | 5e-4 | 280 |

The decay schedule for the learning rate is defined as follows:

$$(q, \Delta t) = \begin{cases} (0.8, 10) & \text{epoch} < 120 \\ (0.7, 10) & \text{epoch} \geq 120 \end{cases}$$

The accuracies' Std is 0.0025.

| A-LeNet5 - d | | | |
|---|---|---|---|
| $\eta$ | $\mu$ | $\alpha$ | epochs |
| 0.032 | 0.92 | 5e-4 | 240 |

The decay schedule for the learning rate is defined as follows:

$$(q, \Delta t) = \begin{cases} (0.8, 10) & \text{epoch} < 120 \\ (0.7, 10) & \text{epoch} \geq 120 \end{cases}$$

The accuracies' Std is 0.0035.

| A-LeNet5 - e | | | |
|---|---|---|---|
| η | μ | α | epochs |
| 0.02 | 0.922 | 1.2e-3 | 240 |

The decay schedule for the learning rate is defined as follows:

$$(q, \Delta t) = \begin{cases} (0.8, 10) & \text{epoch} < 120 \\ (0.7, 10) & \text{epoch} \geq 120 \end{cases}$$

The accuracies' Std is 0.0039.

**Statistics.** Statistics for each architecture were obtained using 10 samples.

**Hardware and software**. We used Google Colab Pro and its available GPUs. We used Pytorch for all the programming processes.